\documentclass[conference]{IEEEtran}
\usepackage[para,online,flushleft]{threeparttable}
\usepackage{soul}
\usepackage{sidecap}
\usepackage{epsfig}
\usepackage{graphicx}
\usepackage{multirow}
\usepackage{hanging}
\usepackage{wrapfig}
\usepackage{subfig}
\usepackage{epstopdf}
\usepackage{float}
\usepackage{booktabs}
\usepackage{algorithm}
\usepackage{algpseudocode}
\usepackage[small,compact]{titlesec}
\usepackage{palatino}
\usepackage[rflt]{floatflt}
\usepackage{subfig}
\usepackage{comment}
\usepackage{amsmath,amssymb}
\usepackage{multirow}
\usepackage{cite}
\usepackage{tabularx}
\usepackage[svgnames]{xcolor}
\usepackage{listings}
\usepackage[T1]{fontenc}
\usepackage[utf8]{inputenc}

\usepackage[english]{babel}
\usepackage[breaklinks,colorlinks,linkcolor=FireBrick,citecolor=Brown,urlcolor=DarkBlue]{hyperref}
\usepackage{url}
\usepackage{color}
\usepackage{lipsum}
\usepackage{titlesec}

\newcommand{\ignore}[1]{}

\titleformat{\paragraph}[runin]{\itshape\bfseries}{}{2pt}{\space}[.]
\titlespacing*{\paragraph}{\parindent}{0pt}{0pt}

\title{Programming Autonomous Machines  \\
    \LARGE Special Session Paper
}

\begin{document}

\author{\IEEEauthorblockN{Shaoshan Liu}
    \IEEEauthorblockA{
        PerceptIn, Fremont, CA, USA \\
        \url{shaoshan.liu@perceptin.io}
    }
    \and
    \IEEEauthorblockN{Xiaoming Li}
    \IEEEauthorblockA{
        University of Delaware, Newark, DE \\
        \url{xli@udel.edu}
    }
    \and
    \IEEEauthorblockN{Tongsheng Geng}
    \IEEEauthorblockA{
        University of California, Irvine, CA, USA\\ 
        \url{tgeng@uci.edu}
    }
    \and
    \IEEEauthorblockN{Stéphane Zuckerman}
    \IEEEauthorblockA{
        Laboratoire ETIS,  UMR 8051, \\ CY Cergy Paris Université, ENSEA, CNRS, \\ F-95000, Cergy, France \\
        \url{stephane.zuckerman@ensea.fr}
    }
    \and
    \IEEEauthorblockN{Jean-Luc Gaudiot}
    \IEEEauthorblockA{
        University of California, Irvine, CA, USA \\ 
        \url{gaudiot@uci.edu}
    }
}

\maketitle

\begin{abstract}
One key technical challenge in the age of autonomous machines is the programming of autonomous machines, which demands the synergy across multiple domains, including fundamental computer science, computer architecture, and robotics, and requires expertise from both academia and industry. This paper discusses the programming theory and practices tied to producing real-life autonomous machines, and covers aspects from high-level concepts down to low-level code generation in the context of specific functional requirements, performance expectation, and implementation constraints of autonomous machines. 
\end{abstract}


\section{Introduction }
\label{sec:intro}
After decades of uninterrupted progress, information technology has so evolved that it can be said we are entering the age of autonomous machines~\cite{liu2022rise}. 
One key technical challenge in this context is the programming of such autonomous machines, as it demands to operate a synergy across multiple domains, including fundamental computer science, computer architecture, and robotics, and requires expertise from both academia and industry. 

This paper discusses the programming theory and practices tied to producing real-life autonomous machines, and covers aspects from high-level concepts down to low-level code generation in the context of specific functional requirements, performance expectation, and implementation constraints of autonomous machines. 

For instance, autonomous vehicles rely on a wealth of specialized components according to the various tasks they are required to perform, which includes (hard) real-time tasks related to the outside environment, \emph{i.e.}, Localization and Navigation, Object Detection and Avoidance, \emph{etc.} Each task communicates its data to other tasks within a strict performance envelop, and may also rely on very different hardware targets, \emph{e.g.}, scheduling with CPUs, GPUs for neural network processing, FPGAs or DSPs for image processing, \emph{etc}~\cite{liu2021pi, liu2017computer, yu2020building, liu2021robotic}.

Thus, there should be expressive ``languages'' to describe, at a high level, what each task should consist of, the appropriate (domain-specific) semantics for it, and at the same time, the interfacing between languages. As a whole, real-time and run-time contexts can freely dictate which part of the underlying hardware will be used to run them at a specific moment during the motion of the target autonomous machine. To do so, simply designing a new set of DSLs is probably necessary, but insufficient. As these autonomous machines tend to heavily rely on Machine Learning techniques for at least some of their tasks, there should also be a way to lower the high-level description---semantic and performance expectation---provided for each task down to an intermediate representation which would allow the compiler to produce code for a family of heterogeneous devices.
These considerations, which are both high and low level, can be unified under a dataflow-oriented hierarchical view of the system.

Hence, the main challenges we wish to address in this paper are related to expressing the tasks of how autonomous machines must perform at high-level, and how the expression is best translated into low-level operations:
\begin{itemize}
    \item How to make autonomous machines more \emph{programmable}, through the use of high-level languages;
    \item How can such high-level languages be translated, \emph{i.e.}, \emph{lowered}, to a suitable set of intermediate representations for a compiler, in order to eventually produce executable code on heterogeneous hardware;
    \item How is the runtime system meant to deal with the heterogeneity of the underlying hardware and dynamic performance envelop, and how can it benefit from both high-level languages, and the knowledge of the low-level machine models it has access to.
\end{itemize}

This paper is organized as follows: 
Section~\ref{sec:amc} explains the computing patterns of autonomous machine computing as well as the programming challenges;
Section~\ref{sec:program} introduces a high-level programming language design to express autonomous machine computing graphs;
Section~\ref{sec:runtime} unveils the methodology and tools we propose to develop the autonomous machine computing system software;
Section~\ref{sec:prototype} shows the experimental results obtained using an early prototype, following some of our methodology; 
Section~\ref{sec:related_work} presents some related work and contrasts our approach with it;
we conclude in Section~\ref{sec:conclusion}.

\section{Autonomous Machine Computing: Problem Statement and Challenges}
\label{sec:amc}
In this section, we introduce the patterns of autonomous machine computing, define the problem statement, and indicate the challenges of programming autonomous machines. In the case of autonomous vehicles, the system is a mission critical real-time system where the system must make “correct” decisions in real-time to react to varying traffic conditions, at a high speed. Multiple computation intensive applications, such as sensor processing, robotic perception, robotic localization, as well as planning and control, etc. are employed to deliver the required high precision tasks. Moreover, the system needs to perform the computation under a strict energy budget for battery considerations~\cite{yu2021designing}. 

\begin{figure*}
    \centering
    \includegraphics[width=0.8\textwidth]{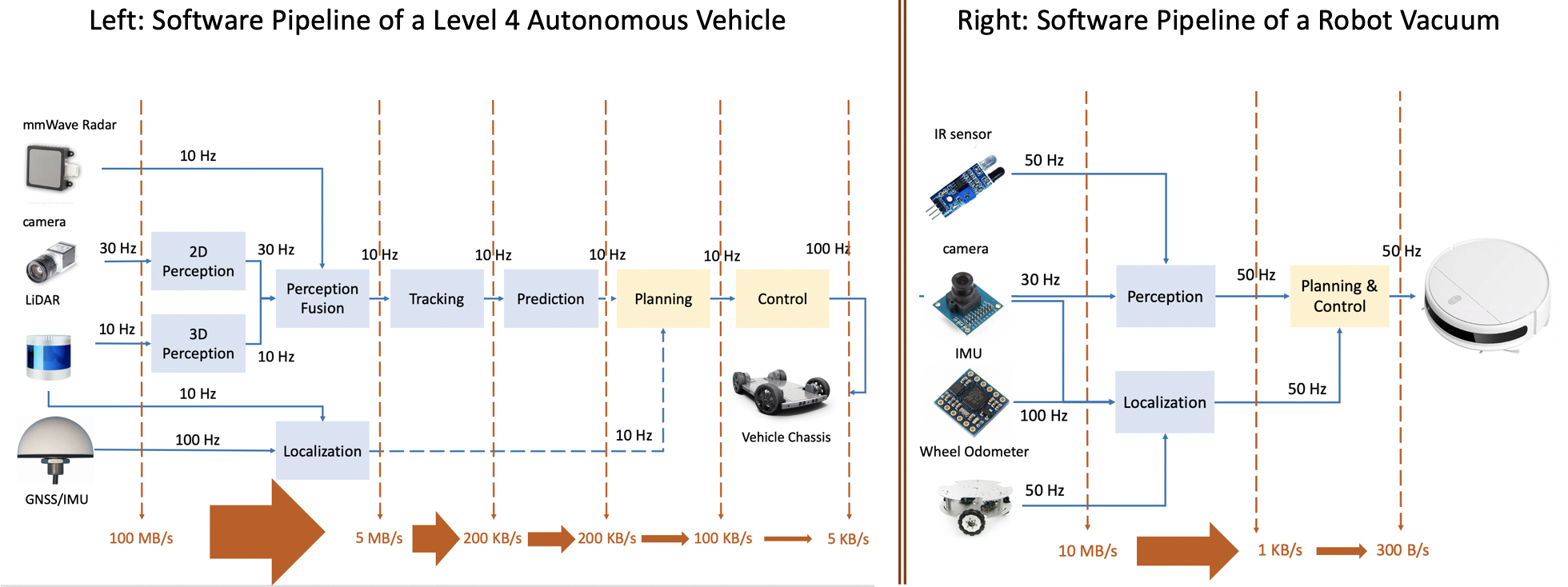}
    \caption{Left: the software pipeline of a level-4 autonomous vehicle. Right: the software pipeline of a robot vacuum. Other software organizations are possible.}
    \label{fig:am}
\end{figure*}

A concrete example is shown in Fig.~\ref{fig:am} a.), which presents an overview of the deep processing pipeline of a commercial level-4 autonomous driving system. Starting from the left side, the sensing system generates raw sensing data from mmWave radars, LiDARs, cameras, Global Navigation Satellite System (GNSS) receivers, and Inertial Measurement Units (IMUs), where each sensor produces raw data at its own frequency. For instance, the cameras capture images at 30 FPS, the LiDARs capture point clouds at 10 FPS, the GNSS/IMUs generate positional updates at 100 Hz. Note that at the sensing stage, an enormous amount of heterogeneous sensing data is generated at a high frequency (\textit{e.g.} over 100 MB/s), the key challenge at this stage is to provide accurate, reliable, and comprehensive information about the physical environment for later processing stages, which can be achieved through precise temporal and spatial synchronization~\cite{liu2021sync}. 

Next, the 2D Perception node (\textit{e.g.} YOLO~\cite{redmon2016you}) consumes raw images for object detection and scene segmentation. the 3D Perception node (\textit{e.g.} PointPillars~\cite{lang2019pointpillars}) consumes raw LiDAR scans for object shape and type detection. The Perception fusion node consumes outputs from the 2D Perception node, the 3D Perception node, as well as the mmWave radars raw data to create a comprehensive perception list of all detected objects. The Localization node (\textit{e.g.} LiDAR odometry and mapping~\cite{zhang2014loam}) consumes and fuses raw LiDAR scans as well as GNSS/IMU data for vehicle positional updates. Hence, the perception and localization system consumes 100 MB/s of raw sensing data and produces 5 MB/s of semantic data in real time. 

The perception list is then fed into the Tracking node at 10 Hz to create a tracking list of all detected objects. The tracking list then is fed into the Prediction node at 10 Hz to create a prediction list of all objects. The Tracking and Prediction system consumes 5 MB/s of perception inputs and further reduces the data size to 200 KB/s. 

At last, both the prediction results and the localization results are fed into the Planning node at 10 Hz to generate a navigation plan. The navigation plan then is fed into the Control node at 10 Hz to generate control commands, which are finally sent to the autonomous machine for execution at 100 Hz. The Control node generates commands with size of 5 KB/s. 

Hence, the autonomous driving system processing pipeline consumes 100 MB/s of raw sensing data at the beginning and generates 5 KB/s of commands at the end. Every 10 ms, the autonomous machine needs to generate a control command. If any upstream node, such as the Perception node, misses the deadline to generate an output, the Control node must still generate a command before the deadline. This could lead to disastrous results as the autonomous machine would then be essentially driving blindly without timely participation from the perception unit. 

In addition to autonomous vehicles, we have identified similar patterns in other autonomous machines, such as the robot vacuum computation graph shown in Fig.~\ref{fig:am} b.), albeit with variations in the depth of the pipeline and the output frequencies. 

Existing real-time computing techniques mainly focus on the understanding of a specific system's real-time behavior through mathematical modeling~\cite{joseph1986finding}, or developing scheduling algorithms for a particular architecture~\cite{liu1973scheduling}. However, as autonomous machine computing is a newly emerging field, neither mature mathematical models to understand the real-time behaviors of autonomous machine systems, nor scheduling algorithms on new architectures (such as the hybrid dataflow and DSA architecture proposed in this article) exist, and an real-time computing challenge for autonomous machines is currently outstanding~\cite{rtss2021challenge}. 

A key observation from our practical experiences of deploying various types of autonomous machines is that \textbf{\textit{autonomous machine computing can be expressed as dataflow graphs~\cite{liu2021dataflow}}}. Indeed, as exemplified by ROS~\cite{quigley2009ros}, autonomous machine application programmers usually connect a limited set of basic operators together to form a computation graph. ROS is a great start, but in a way ROS is very heavy and the learning curve for ROS can be steep. Autonomous machine application programmers have to learn many different forms of communication between the ROS nodes and handle the communication and resource allocation details themselves. Many of these details are not directly related to the functionality of the autonomous machine under development. \textbf{\textit{Hence, a language is required to allow autonomous machine application programmers to express their dataflow graphs in a very concise and precise way.}}

Second, ROS exposes a lot of low-level details to autonomous machine application programmers. Unfortunately, most autonomous machine application programmers don’t have the expertise on optimizing for performance or energy efficiency~\cite{liu2021engineering}, hence often leading to hacks that make autonomous machine application programs un-maintainable, or programs that are not performant. To exacerbate the problem,  the computer architecture and design automation communities often come up with hardware accelerators trying to help autonomous machine application programmers. But as more and more accelerators come into existence, autonomous machine application programmers are not able to manage the complexity and they are not building autonomous machine SoCs to integrate these accelerators.  

Therefore, we need a autonomous machine development stack, likely including language, compiler and runtime, to allow autonomous machine application programmers focusing on application programming instead of system engineering. Based on the performance and energy consumption specifications from the autonomous machine application programmers, \textbf{\textit{the runtime should be able to automatically dispatch operators to the underlying compute substrate (CPU, GPU, DSP, accelerators), or simply refuses to execute the program if the specifications cannot be met.}}

\begin{figure}
    \centering
    \includegraphics[width=0.8\linewidth]{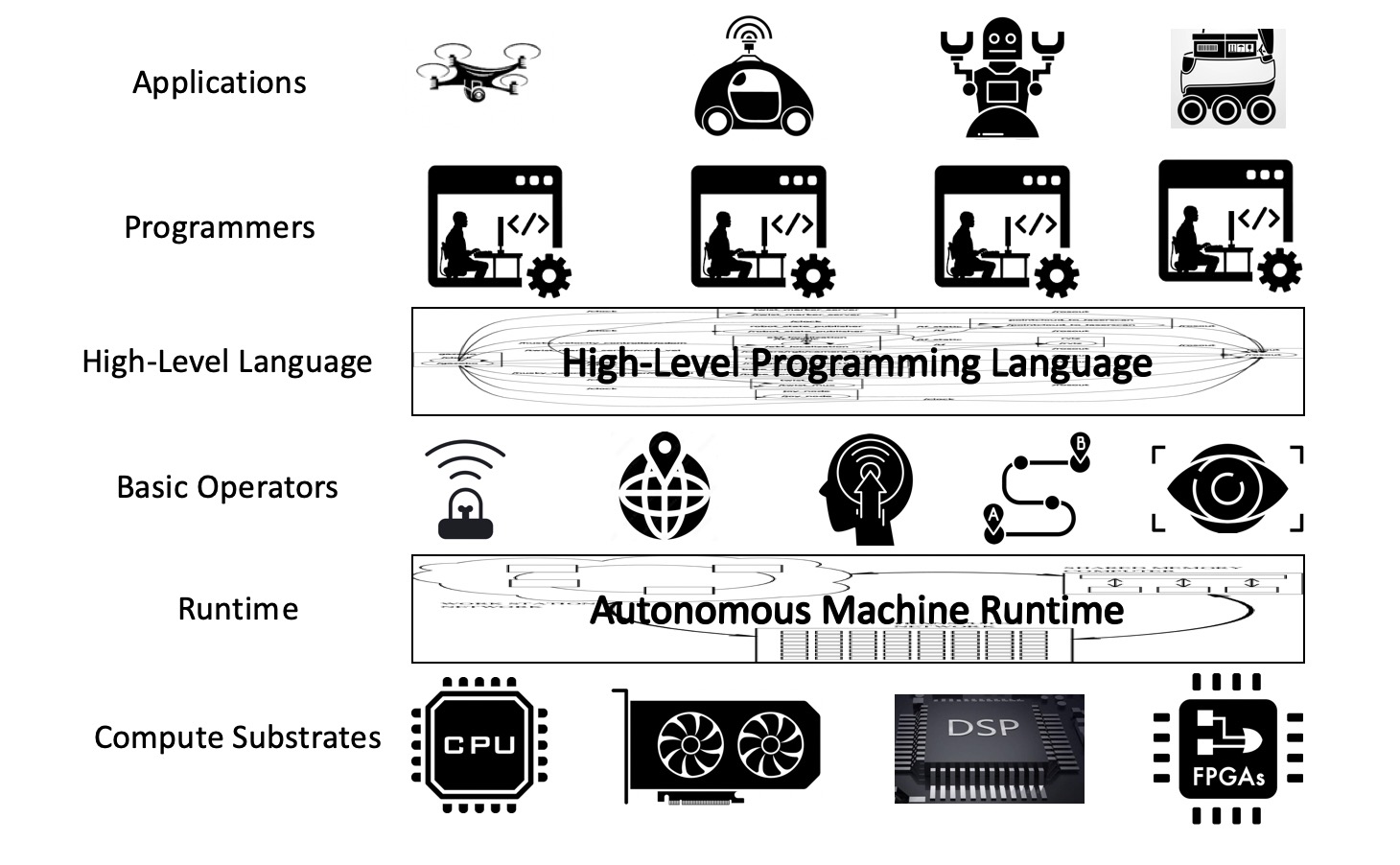}
    \caption{Computing Stack for Autonomous Machines}
    \label{fig:stack}
\end{figure}

We summarize our proposal in Fig.~\ref{fig:stack}, the high-level language allows the application programmers focusing on autonomous machine application programming instead of the low-level system details, whereas the autonomous machine runtime automatically judges whether a proposed autonomous machine computing graph can be mapped to the compute system, and automatically maps individual operators to the best-suited compute substrates.  

\section{High Level Programming}
\label{sec:program}
\begin{algorithm*}[thb]
\caption{$\lambda$-calculus expression of the computation graph shown in Fig.~\ref{fig:am}b.)}
\label{alg:rv}
\begin{algorithmic}[1]
\Require $IR (frequency >= 50 Hz)$ \Comment{infrared sensor}
\Require $Camera (resolution = 320\times240; frequency >= 30 Hz)$  
\Require $IMU (frequency >= 100 Hz)$ 
\Require $WO (frequency >= 50 Hz)$  \Comment{wheel odometry sensor}
\Require $2DPerception (frequency >= 50 Hz)$ 
\Require $Localization (frequency >= 50 Hz)$ 
\Require $Control (frequency >= 50 Hz)$ 
\State $perc \gets \lambda(IR)\lambda(Camera)\lambda(2DPerception)[2DPerception(IR,Camera)]$
\State $loc \gets \lambda(Camera)\lambda(IMU)\lambda(WO)\lambda(Localization)[Localization(Camera,IMU,WO)]$
\State $cmd \gets \lambda(perc)\lambda(loc)\lambda(Control)[Control(perc,loc)]$
\end{algorithmic}
\end{algorithm*}
In this section, we introduce how a concise and precise high-level language can be developed to represent the computation graphs of autonomous machines.  As autonomous machine computing can be represented as dataflow graphs, naturally a functional programming paradigm provides an efficient way to describe the behavior of autonomous machines. Indeed, in addition to being a suitable programming model for dataflow architectures, functional programming has already been widely used in autonomous machines, see the~\nameref{para:related_work_dsl} paragraph in section~\ref{sec:related_work}.
With functional programming, autonomous machine application programmers can describe their applications with a few lines of specifications and code as illustrated in the following examples.

Algorithm \ref{alg:rv} shows a $\lambda$-calculus expression ~\cite{barendregt1984lambda} describing the cleaning robot computation graphs shown in Fig.~\ref{fig:am}. Note that the first few lines of \textit{Require} statements specify the configurations (\textit{e.g.} image resolution) and the performance specifications for the compute substrates. The proposed compiler goes through these statements and ensures that these performance requirements can be met. 
The next few lines of code specify the \textit{Computation Graph}. Based on the performance requirements, i.e., the \textit{Require} and \textit{Graph}, the compiler will allocate the corresponding buffer space to facilitate communications between the nodes to avoid timing violations caused by excessive stalls. 
Note that with this approach, complex autonomous machine applications can be expressed with a few lines of code using functional, event-driven programming.

Similarly, Algorithm \ref{alg:av} describes the L4 autonomous vehicle computation graph shown in Fig.~\ref{fig:am}, a more complex computation graph compared to the robot vacuum. This example employs more sensors, and constructs a dataflow graph that is much deeper and wider. Nonetheless, similar to the robot vacuum case, the L4 autonomous vehicle dataflow graph can also be described with only a few lines of \textit{Require} as well as \textit{Computation Graph} statements. 

\begin{algorithm*}
\caption{$\lambda$-calculus expression of the computation graph shown in Fig.~\ref{fig:am}a.)}
\label{alg:av}
\begin{algorithmic}[1]
\Require $Radar (frequency >= 10 Hz)$
\Require $Camera (resolution = 1920X1080; frequency >= 30 Hz)$  
\Require $LiDAR (resolution = 64 beams; frequency >= 10 Hz)$ 
\Require $GNSS (frequency >= 100 Hz)$
\Require $2DPerception (frequency >= 30 Hz)$ 
\Require $3DPerception (frequency >= 10 Hz)$ 
\Require $PerceptionFusion (frequency >= 10 Hz)$ 
\Require $Localization (frequency >= 10 Hz)$ 
\Require $Tracking (frequency >= 10 Hz)$
\Require $Prediction (frequency >= 10 Hz)$ 
\Require $Planning (frequency >= 10 Hz)$ 
\Require $Control (frequency >= 100 Hz)$ 
\State $2dperc \gets \lambda(Camera)\lambda(2DPerception)[2DPerception(Camera)]$
\State $3dperc \gets \lambda(LiDAR)\lambda(3DPerception)[3DPerception(LiDAR)]$
\State $loc \gets \lambda(LiDAR)\lambda(GNSS)\lambda(Localization)[Localization(LiDAR,GNSS)]$
\State $percfus \gets \lambda(2dperc)\lambda(3dperc)\lambda(PerceptionFusion)[PerceptionFusion(2dperc,3dperc)]$
\State $traj \gets \lambda(percfus)\lambda(Tracking)[Tracking(percfus)]$
\State $pred \gets \lambda(traj)\lambda(Prediction)[Prediction(traj)]$
\State $plan \gets lanning(pred,loc)]$
\State $cmd \gets \lambda(plan)\lambda(Control)[Control(plan)]$
\end{algorithmic}
\end{algorithm*}

To realize the promise of this approach, the key is to rely on the system software (See Section~\ref{sec:runtime}) to hide the details tied to real-time constraints in the underlying compute substrates, by providing time specifications to meet deadlines. 

The system software must retain enough semantic information to be able to correctly map specific computations to the right hardware compute unit, and equally importantly, according to real-time constraints. The details regarding the target compiler \& runtime system are given in Section~\ref{sec:runtime}. 


\section{System Software: Compiler and Runtime}
\label{sec:runtime}
Leveraging functional constructs naturally fit in our dataflow-inspired context, as shown in the previous section. Such a high-level language  could be one that implements Algorithms~\ref{alg:rv} and~\ref{alg:av} as part of a declarative syntax, or as part of an imperative syntax with added functional mechanisms
\footnote{Several languages take this path: OCaml, Scala, and more recently Swift, Rust, \emph{etc.}}. 
In general, type safety, immutable states, and functional constructs tend to greatly help toward data-race free parallel programs ~\cite{hughes1989functional,wadler1992essence,peyton1993imperative,hudak1989conception}. 

\subsubsection{A Compiler for Autonomous Machines}
\label{sec:runtime:compiler}

As explained in Section~\ref{sec:program}, the right high-level syntax helps the programmer tremendously with productivity. 
However, the compiler still requires explicit machine-specific information, called annotations. These \emph{annotations} will provide resource mapping information, real-time deadline and dataflow dependencies, as well as declare when a ``basic computation block'' should be used (through the use of \emph{intrinsic} functions).


\paragraph{Expressing Resource Mapping Preferences}~
\emph{Requirements} (See Algorithm~\ref{alg:rv} and~\ref{alg:av}) will be used to express mandatory compute-to-device mapping. They will most likely be used when the programmer already has profiling information (\emph{e.g.}, through micro-benchmarking) at their disposal. 
In addition, \emph{hints} are also available. They are ``advice'' to express resource mapping preferences which are useful when no precise timing information about a given computation is already known, but there is general information which i, \emph{e.g.}, control-intensive code tends to best run on general-purpose CPUs, whereas image processing kernels tend to behave very well on GPUs or DSPs. 
Once processed, annotations will be passed from the compiler down to the runtime during code generation (see Section~\ref{sec:runtime:rts} below).

\paragraph{Expressing Autonomous Machines Intrinsics}~
The compiler should also know about autonomous machines ``basic computation blocks,'' expressed as \emph{intrinsic functions}. \emph{Intrinsics} represent a processing step often found in autonomous machines programs, \emph{e.g.}, feature extraction on a stream of pixels. The compiler can then substitute the intrinsic call with library calls, or code generated and optimized for the hardware target. 

\paragraph{Expressing Real-time Constraints}~
Autonomous machines must obey real-time (RT) constraints for safety and security reasons. 
The compiler must accept and enforce RT constraints annotations, as with resource mapping. 
Moreover, in addition to time constraints, the compiler must accept computation and/or data dependence descriptions, as this will help it compute the overall RT-based deadlines and make all computation blocks fit the individual deadlines expressed by the programmer. This will naturally result in a graph-based representation of RT constraints analyzable by the compiler. 


In any case, the programmer should not have to directly deal with the low-level runtime API to express computation-to-hardware mapping, and the tedious tuning to satisfy performance requirements. Instead, the compiler should be able to convert the programmer's source code into either direct CPU/GPU/DSP/FPGA code when necessary, or call to an already-optimized library if it is available on the target system\footnote{For instance, Intel's compiler performs pattern-matching in the code to substitute naïve linear algebra kernels with MKL ones.}.

Moreover, the use of different hardware resources should lead to the static allocation of low-level buffers to allow efficient communication between compute devices, but also flexible/dynamic memory (buffer) management to deal with the production and consumption of data between computation blocks. The latter buffers will see their size vary dynamically according to the real-time context.


\subsubsection{Associated Runtime System}
\label{sec:runtime:rts}

The high-level syntax allows low time-to-solution programs generation for autonomous machines; the compiler will provide in-depth program analysis and, with the help of the programmer's annotations, will also express ``good computation-to-device mapping defaults'' to the runtime system, including memory management guidelines. 


Such a runtime must fulfill several objectives. It must: (1) Of course, be low-overhead; (2) be able to map computation tasks to the best fit available compute unit at the time of scheduling; (3) fulfill real-time requirements in a very dynamic context; (4) be able to deal with real-time based task dependencies.

\paragraph{Low resource management overhead}~ 
Autonomous machines include \emph{e.g.}, self-driving vehicles, which may be running at high speeds, and require to meet hard RT deadlines. It is thus paramount that task and resource scheduling takes as little time as possible.

\paragraph{Efficient computation-to-device mapping}~ 
The target hardware composing autonomous machines is highly heterogeneous, and can be comprised of, \emph{e.g.}, CPUs, GPUs, DSPs, FPGAs, \emph{etc.} 
As a result, the efficiency in terms of performance, and/or power and energy consumption may vary a great deal between two similar-yet-different hardware targets. 

The more challenging task is the dealing with when the running application may drastically change its behavior when environmental constraints make it necessary. Hence, the runtime must be able to dynamically adapt its resource management policy to accommodate new constraints. 
Hence, the runtime system must dynamically adapt computation-to-resource mapping, as the situation evolves. In the absence of precise timing information, the compiler must rely on the ``good defaults information'' provided by the compiler.

\paragraph{Real-time constraints satisfaction}~
RT constraints will be expressed as meta-data fields in the runtime, to be taken into account by the resource manager~\cite{liu2021rtss}. The latter will then evaluate which mapping policy is the best at the time of scheduling. The policy itself should obey the \emph{hint} vs. \emph{requirement} annotations expressed by the programmer, \emph{i.e.}, and compute unit specific \emph{requirements}.

However, both the compiler and the runtime system should be able to warn against impossible schedules when RT constraints will be obviously violated. Taking RT constraints into account will naturally result in the runtime building a semi-static data-driven task dependence graph. This will in turn lead to the use of established scheduling policies, both static (\emph{e.g.}, HEFT~\cite{TopcuogluHaWu2001} and its variants), and more dynamic ones (\emph{e.g.}, PDAWL~\cite{geng2022profile}).


\subsubsection{Bridging the Compile-Runtime System Gap} 
\label{sec:runtime:mlir}

So far, it may seem the goals expressed for both the compiler and the runtime system may match, but that they may be quite hard to realize in practice from an implementation perspective: how do we make the needs expressed in a high-level language w.r.t. hardware resource usage, real-time, data, and code constraints, and using high-level concepts to eventually perform low-level optimized computations work together?

A promising lead is the use of hierarchical and orthogonal intermediate representation \emph{dialects}. The use of a compiler which allows for progressive lowering or even raising of the required properties could help tremendously. MLIR~\cite{LattnerEtAl2020,LattnerEtAl2021} allows a compiler writer to specify a dialect, which can also be combined with other existing ones, in order to provide domain-specific abstract syntax trees (ASTs) which retain the right level of information to make informed decisions, such as choosing which version of a given computation kernel should be used, depending on the available devices on the target platform, the hints or guidelines provided by the programmer, \emph{etc.}. The end-result can be lowered down to LLVM's intermediate representation, which can then be used to generate actual code. 

MLIR also allows for outlining regions of codes, \emph{i.e.}, transforming the currently analyzed code into a function call that can then be ``injected'' into the original code. Hence, a language intrinsic may call for a dot product computation. Such operations can be defined in MLIR, such that the compiler ``knows'' about them, and has been informed of their properties (which in turn may help with uncovering optimization opportunities). Hence, if an optimized \texttt{sdot} code already exists for the target platform, MLIR can outline the intrinsic call to insert a call to the real function.

\section{Prototype: Description and Results}
\label{sec:prototype}
In this section, we utilize the ORB~\cite{rublee2011orb} as an example to illustrate the programming methodology we presented in the previous section. 
ORB is an essential component in Simultaneous Localization And Mapping (SLAM)~\cite{durrant2006simultaneous}, 3D reconstruction~\cite{geiger2011stereoscan}, and many other autonomous applications~\cite{mur2015orb}. 

\begin{figure}
    \centering
    \includegraphics[width=1.\linewidth]{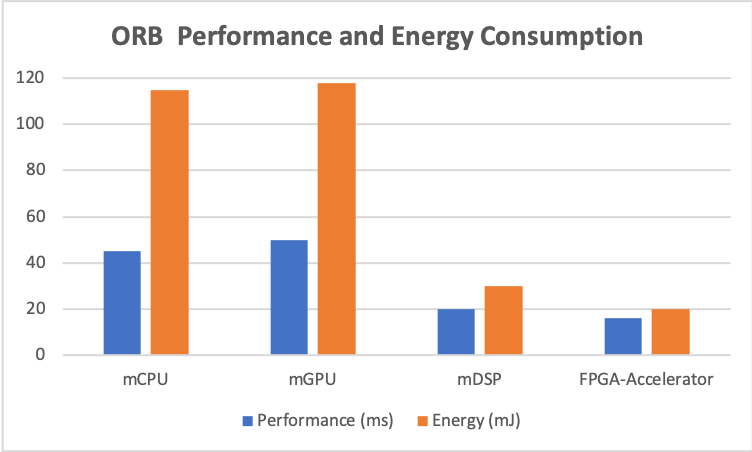}
    \caption{ORB performance and energy consumption on mobile CPU, mobile GPU, mobile DSP, and FPGA-based accelerator~\cite{fang2017fpga}}
    \label{fig:perf}
    \vspace{-15pt}
\end{figure}

We have implemented SLAM on a quad-core ARM v8 mobile SoC and identified that the feature extraction stage (in ORB) consumes >50\% of the CPU resources~\cite{tang2018pi}. Further, we have run ORB on a Qualcomm Snapdragon 820 SoC and on a FPGA-based accelerator~\cite{fang2017fpga}, the performance and energy consumption profiling results are shown in Fig.~\ref{fig:perf}. It shows a highly diverse performance portfolio even for the relatively simple ORB workload.

Per discussion before, the main challenge faced by the industry is that most autonomous machine engineers don't have the skills to optimize the autonomous machine system for best performance or energy efficiency. Hence it is imperative to have a compiler and  runtime system to automatically map the autonomous machine basic blocks onto the appropriate compute substrates.  An early prototype of a runtime system for autonomous machine vision workloads was proposed by Liu \emph{et al.}~\cite{liu2021pi}.

In this case study, we will demonstrate the capability of the proposed programming language for ORB in autonomous machine applications from three aspects: the capability to specify performance expectation/contract in programming; the enforcement of the contracts in the system software through compilation techniques, runtime adaptation, and dynamic mapping to computing substrates. We will evaluate the capability through an experiment that simulates the use ORB in autonomous machines and with multiple data benchmarks.

\subsection{Performance Contracts and Envelop}

Performance contracts for autonomous machines needs to provide specific guidelines. The meaningful timing-related metrics include latency/frequency, throughput, variance, \textit{etc.} Also, the power consumption (Watts) and the energy consumption (Joules) may also need to be specified.

However, some performance contracts may not feasible. The proposed programming language compilation and execution support can help the specification of performance contracts, and equally important, can guide developers establish the feasible performance ``envelop'' based on computing substrates as well as characteristic of input data.  



\subsection{Compiler's and Runtime's Enforcement of Performance Contracts}

The system software for the proposed language will enforce the performance contracts. The main techniques includes contract decomposition, code versioning, and profiling and adaptation. 

``Contract decomposition'' means that the performance requirements for a higher level program construct, such as library modules, are broken down into subcontracts for composing program constructs. Usually it is only advisable for engineers to specify those top-level constructs and the language compiler and runtime system will automate the process of the contract decomposition.

``Code versioning'' means that multiple versions of a program construct might be generated to adapt to different performance requirements or computing substrates. For example, ORB might need to be optimized in different ways when it is to run at 1000Hz in contrast to running at 100Hz, or is to run onto different hardware.

``Profiling and adaptation'' is the methodology that profiles the performance, gauges the deviation from the performance contract, and remedies the deviation.

In this case study, we will demonstrate an initial capability of our language to decompose higher-level performance contract for ORB, then generate multiple versions to adapt to different scenarios, and at runtime, automatically deal with performance deviations.

\subsection{Experiment Setup and Evaluation}

In this study, we simulate the ORB~\cite{rublee2011orb} from OpenCV~4
 in our language. Different from the implementation as illustrated in Fig.~\ref{fig:perf} that is based on ad-hoc tuning, the prototype is re-implemented with the core concepts proposed in this paper, and runs in a computer system with heterogeneous computing resources with dynamically changing image inputs. In this process, the system software support will be able to automatically establish the performance envelop of various hardware + input scenarios, and at runtime, maintain pre-specified performance contracts. 

Specifically, the hardware setup is as follows: the CPU is AMD Ryzen 5950X with 16 cores running at 3.4 GHz; and the GPU is a Nvidia GeForce RTX 3080 with 8960 CUDA cores running at 1.71 GHz. The data benchmarks we use are KITTI~\cite{geiger2013vision} and TUM RGB-D~\cite{sturm2012benchmark} datasets.

\subsubsection{Experiment 1: Establish the Performance Envelop}

In order to accept and enforce user-specified performance contracts through the whole development pipeline, our proposed system needs to first answer the question of what are the practical performance envelops of a piece of code on target hardware. The performance metrics that compose a performance envelop typically include latency, throughput and variability. The metrics may leading to conflicting compilation tuning choices, \emph{e.g.}, optimizing for latency may lead to less throughput. Therefore, the performance envelop is actually a compromise, \emph{i.e.}, a Pareto curve of local optima.

\begin{figure}[htb]
  \centering
    \hspace*{-.5cm}
    \includegraphics[width=0.8\linewidth]{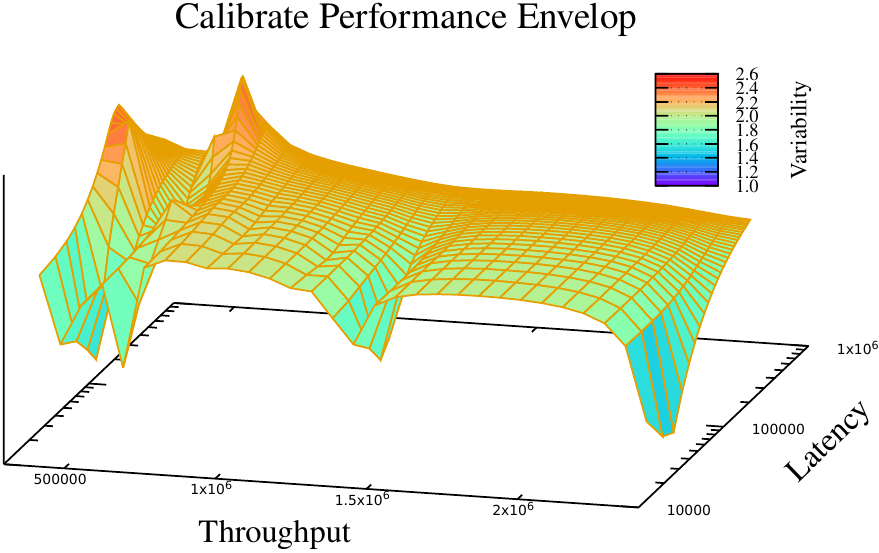}
  \caption{Performance Envelop of ORB+Matching}
  \label{fig:performance_envelop}
  \vspace{-10pt}
\end{figure}

In this experiment, we demonstrate a collection of preliminary compiler and profiling techniques to automatically establish the performance envelop of ORB on the target hardware. Figure~\ref{fig:performance_envelop} illustrates the performance envelop as a Pareto surface in the 3D space of latency, throughput and variability.  The results confirm that the same code, in this case ORB, can be implemented and tuned towards widely different performance contracts. Our system can map the envelop of possibilities and help developers select a Pareto-optimal point that suits their specific goals.


\subsubsection{Experiment 2: Decompose and Dynamically Maintain Performance Contracts}

To enforce a user-provided overall performance contract, the development system needs to do two things: breaking down the overall contract onto
sub-components, and adapting to runtime deviations from nominal. These two tasks are probably the most involving and challenging problems for autonomous machine engineers. In this experiment, we will demonstrate with ORB how our proposed system can automate the solutions for these two tasks.

\begin{figure}[htb]
  \centering
    \includegraphics[width=\linewidth]{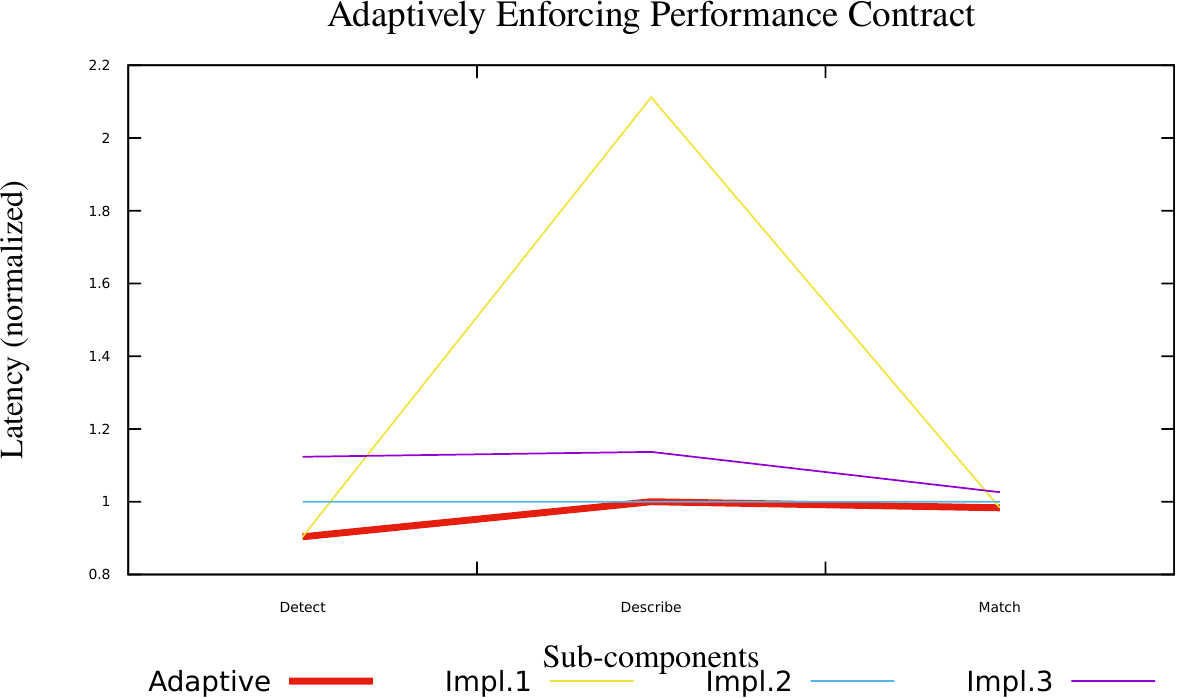} 
  \caption{Performance Envelop of ORB+Matching}
  \label{fig:performance_decomposition}
  \vspace{-10pt}
\end{figure}

A typical working flow for ORB has three steps: key-points finding, descriptor generation, and point matching. Here, we show that the performance contract for ORB can be broken down into the performance expectations on these sub-components, and at runtime, we can dynamically adapt the implementation choices to maintain an overall contract. Figure~\ref{fig:performance_decomposition} demonstrates different strategies and dynamism lead to different performance at these stages. Each line represents a different implementation strategy that involves the tuning of resource utilization, devices and optimization prioritization. The "Adaptive" lines shows that our system can use input- and runtime-derived heuristics to adapt. The results show that our proposed system can adapt and maintain the given performance contract despite the dynamism from input scenarios or execution situations.


\section{Related Work}
\label{sec:related_work}

In this section, we review related work in autonomous machine programming. 

\noindent \paragraph{Adaptive Dynamic Runtime Frameworks}~
\label{para:related_work_runtime}
Autonomous Machines (AM) need to operate across a variety of challenging environments which impose  continuously-varying deadline and runtime-accuracy tradeoffs on the computing pipelines~\cite{yu2020building,lin2018architectural,sung2022brief,liu2022brief}. To meet stringent  safety-critical requirements  which ensure the fulfillment of strict deadlines in worst-case scenarios in real time, optimized computationally intensive AM algorithms, combined with sophisticated scheduling techniques are implemented onto (heterogeneous) multi-core systems~\cite{bradski2000opencv, campos2021orb, gog2022d3,redmon2017yolo9000}, hardware accelerators (ASIC~\cite{li2019879gops,suleiman2019navion}, FPGA~\cite{fang2017fpga,asgari2020pisces,gautier2019fpga,liu2019eslam,liu2021archytas,gan2021eudoxus}), mobile SoC~\cite{klein2009parallel,weimert2010natural}, \textit{etc.} However, the commuting resources under-utilization issue is a pervasive problem  since the applications in different pipeline stages exhibit environment-dependent runtime features which exhibit a wide variance between average and worst-case 
runtimes~\cite{yu2020building,lin2018architectural,gog2022d3,alcon2020timing}. 
Thus, to efficiently execute AM algorithms and maximize runtime-accuracy, the framework must interact with a continuously-evolving environment~\cite{gog2022d3,alcon2020timing}.

Adaptive real-time systems~\cite{rosu1997adaptive,block2008adaptive,geng2022profile} determine the best service at runtime from multiple application-defined levels using feedback-based scheduling algorithms and optimization techniques. Gan \textit{et al.}~\cite{gan2020eudoxus} proposed Eudoxus, a framework for autonomous machine localization, to adapt to different operating scenarios by fusing fundamental algorithmic primitives. Liu \textit{et al.}~\cite{liu2021archytas} proposed Archytas, a framework that automatically generates synthesizable localization accelerators from high-level algorithm descriptions given power, latency, and resource specifications. 
Liu \textit{et al.}~\cite{liu2021pi} proposed $\pi$-RT, a robotic vision runtime framework that dynamically manages task executions onto local mobile SoC systems with multiple accelerators, as well as on the cloud considering latency, throughput and energy constraints. 
These systems inspire elements of our graph-based deadline-sensitive runtime framework design equipping with flexibility of graph features to deploy tasks based on the tasks features, the  dynamic real-time constraints as well as the high-performance requirements.  


Gog \textit{et al.}~\cite{gog2022d3} proposed their D3 (Dynamic Deadline-Driven) execution model, which decomposes the AM applications as fine-grain computation graphs along with an environment-based deadline policy and centralizes the management of deadlines, and ERDC, which utilizes proactive strategies and exception handlers to  adaptively execute the fine-grained events and handle deadline misses cases, and finally to fulfill varying deadlines demanded by the environment. Similar to D3/ERDC, to obtain a good runtime-accuracy tradeoff, an anytime planning algorithm~\cite{karaman2011sampling,xu2012real} and multiple task execution versions are utilized, which can guarantee that as least one event can be completed before deadline or can release a coarse-grained event to downstream, if the current event deadline is missing, to make sure the overall deadlines are satisfied. 

The most notable difference between D3/ERDC and our approach is that our envisioned runtime framework organizes AM tasks in a graph structure and targets on platforms with various and plug-in-enable (changeable) hardware resources. Furthermore, user-friendly APIs will be provided to deal with specific user-defined constraints.

\paragraph{Robotics Domain-Specific Language (DSL) Models} 
~\label{para:related_work_dsl}
Functional programming (FP)~\cite{hughes1989functional,wadler1992essence,peyton1993imperative,hudak1989conception} provides a high-level and declarative way which has been used in analyzing, describing and verifying the behaviors of AMs/robotics. Peterson \textit{et al.}~\cite{peterson1999lambda} developed Frob, a domain-specific language embedded in Haskell~\cite{peterson1997haskell} for robot control, which describes the interaction between a robot and its stimuli in a purely functional manner. Frob  hides the details of low-level robot operations, promotes a style of programming largely independent of the underlying hardware and supports complex control regiments in a concise and reusable manner. Hudak \textit{et al.}~\cite{hudak2002arrows} proposed Yampa, a DSL to be utilized to program industrial-strength hybrid mobile robots~\cite{nilsson2002functional,pembeci2002functional} with real-time constraints. This approach demonstrates how functional programming can ease the task of the robotics application programmer compared to imperative programming models.
The primary aspects of FP are lamdbas, map operations, and recursion. Most modern languages now support all three of those aspects. In addition, multiple multi-paradigm languages have integrated functional aspects to their core, \emph{e.g.}, JavaScript, C\#, Rust, \emph{etc.}

Low-level wise, open-source C/C++ libraries  such as OpenCV~\cite{bradski2000opencv}, DBoW2~\cite{galvez2012bags}, g2o~\cite{kummerle2011g} \textit{etc.}, are widely utilized in AM algorithms. These libraries can be implemented onto multiple hardware platforms/accelerators. Furthermore, assisted with High-Level Synthesis (HLS) tools~\cite{nane2015survey}, C/C++/Python functions can be compiled into logic elements and layout onto FPGA. 

On the other side, MLIR, see section~\ref{sec:runtime:mlir}, has been integrated into the Tensorflow and PyTorch ecosystem~\cite{pienaar2020mlir}, Deep Learning frameworks~\cite{sommer2022spnc,neuendorffer2021evolution,martinez2022hdnn}, High-level synthesis (HLS) tools ~\cite{urbach2022hls,ye2021scalehls}, \textit{etc.}, that affects a lot of AM algorithms optimization/ implementation, and provides a good interface between the AM programming language development and application implementation.

\section{Conclusion}
\label{sec:conclusion}
This paper presents a vision for a holistic software stack design to efficiently and productively program autonomous machines. The stack is divided into three main parts: a declarative high-level language geared specifically toward autonomous machines; an optimizing compiler which can be driven or at least guided by the programmer to pick the best computational kernels and efficiently map them to the target heterogeneous hardware; a runtime system of which the compiler is aware, capable of handling hardware constraints as well as real-time induced tasks and data dependencies. The stack in turn exposes the need to provide a set of performance contracts between the programmer, the compiler, and the runtime system, for a given hardware target. A case study centered around ORB and image matching is tackled, to provide a view of what the system software's performance contract could look like, leveraging performance measurements of actual runs on a general-purpose multicore + GPU combination. We believe the proposed design provides a precise and concise abstraction of the underlying autonomous machine computing system, which greatly improves the productivity of autonomous machine programming.  

\bibliographystyle{ieeetr}
\bibliography{bib/auto.bib, bib/others.bib}

\end{document}